\documentclass{article}[12pt]
\usepackage{fullpage}
\usepackage{amsfonts}
\usepackage{amsmath}
\usepackage{amsthm}
\usepackage{graphicx}

\usepackage{amsmath,bm}

\newtheorem{lemma}{Lemma}

\newtheorem{theorem}[lemma]{Theorem}
\newtheorem{corollary}[lemma]{Corollary}
\newtheorem{definition}{Definition}

\newcommand{\newmcommand}[2]{\newcommand{#1}{{\ifmmode {#2}\else\mbox{${#2}$}\fi}}}
\newcommand{\newmcommandi}[2]{\newcommand{#1}[1]{{\ifmmode {#2}\else\mbox{${#2}$}\fi}}}
\newcommand{\newmcommandii}[2]{\newcommand{#1}[2]{{\ifmmode {#2}\else\mbox{${#2}$}\fi}}}
\newcommand{\newmcommandiii}[2]{\newcommand{#1}[3]{{\ifmmode {#2}\else\mbox{${#2}$}\fi}}}

\newcommand{\E}[2]{{\bf E}_{#1}\left[ #2 \right]}
\newcommand{\D}{{\cal D}}

\renewcommand{\P}[2]{{\bf P}_{#1}\left[ #2 \right]}
\newcommand{\reals}{\mathbb{R}}

\newcommand{\SP}[1]{{\cal P}^{#1}} 
\newcommand{\R}{R}      

\newcommand{\state}{{\bf \Psi}}

\newcommand{\pot}{\phi}
\newcommand{\potPQ}{\pot_{\learnerM,\adversM}}

\newcommand{\finalPot}[1]{\pmb{\phi}_{#1}}
\newcommand{\finalPotT}{\finalPot{T}}
\newcommand{\finalPotR}{\finalPot{\realT}}
\newcommand{\finalPotTau}{\finalPot{\tau}}
\newcommand{\upperpot}{\pot_{\learnerM}^{\downarrow}}
\newcommand{\upperpotb}{\pot_{\learnerMb}^{\downarrow}}
\newcommand{\upperpotd}{\pot_{\learnerMd}^{\downarrow}}
\newcommand{\upperpotMdk}{\pot_{\learnerMdk}^{\downarrow}}
\newcommand{\upperpotMdj}{\pot_{\learnerMdj}^{\downarrow}}

\newcommand{\lowerpot}{\pot_{\adversM}^{\uparrow}}
\newcommand{\lowerpotb}{\pot_{\adversMb}^{\uparrow}}
\newcommand{\lowerpotd}{\pot_{\adversMd}^{\uparrow}}

\newcommand{\lowerpotMdk}{\pot_{\adversMdk}^{\uparrow}}

\newcommand{\lowerpotMdkpar}[1]{\pot_{\adversMdkpar{#1}}^{\uparrow}}

\newcommand{\realT}{\mathcal{T}}  

\newcommand{\Ilat}[1]{\pmb{I}_{#1}}
\newcommand{\Klat}[1]{\pmb{K}_{#1}}
\newcommand{\score}{\Phi}
\newcommand{\upperscore}[1]{\score_{#1}^{\downarrow}}
\newcommand{\lowerscore}[1]{\score_{#1}^{\uparrow}}

\newcommand{\upperscoreM}{\upperscore{\learnerM}}

\newcommand{\lowerscoreM}{\lowerscore{\adversM}}
\newcommand{\lowerscoreMd}{\lowerscore{\adversMd}}
\newcommand{\learnerM}{P}
\newcommand{\learnerMb}{P_I}
\newcommand{\learnerMd}{P_D}
\renewcommand{\l}{\learnerM}
\newcommand{\learnerMdk}{P_{D(k)}}
\newcommand{\learnerMdj}{P_{D(j)}}

\newcommand{\adversM}{Q}
\newcommand{\adversMb}{Q_I}  
\newcommand{\adversMd}{Q_D}  
\newcommand{\adversMdkpar}[1]{Q_{D(#1)}}
\newcommand{\adversMdk}{\adversMdkpar{k}}  

\newcommand{\Bias}{B}
\newcommand{\deltat}{\Delta t}

\newcommand{\Binom}{\mathbb{B}}

\newcommand{\var}{\mbox{Var}}
\newcommand{\V}{V}

\newcommand{\at}[1]{\left\{ \left. #1
    \right|_{\begin{tiny}\begin{matrix}
          \tau,\rho=\\t_i,\R \end{matrix} \end{tiny}}
    \pot(\tau,\rho)\right\}}
\newcommand{\att}[1]{\left\{ \left. #1  
\right|_{\begin{tiny}\begin{matrix} \tau,\rho=\\t_i+g \deltat_i,\R_i+g
      r_i \end{matrix} \end{tiny}}
\pot(\tau, \rho)\right\}}

\newcommand{\atI}[1]{\left\{ \left. #1  
\right|_{\begin{tiny}\begin{matrix}
      x,y=\\x_0,y_0 \end{matrix} \end{tiny}}
f(x,y) \right\}}
\newcommand{\atII}[1]{\left\{ \left. #1
\right|_{\begin{tiny}\begin{matrix}
      x,y=\\x_0+t\Dx,y_0+t\Dy \end{matrix} \end{tiny}}
f(x,y) \right\}}

\newmcommandi{\paren}{\left({#1}\right)}
\newmcommandi{\brac}{\left[{#1}\right]}

\title{Optimal online Learning using potential functions}
\author{Yoav Freund}
\begin{document}

\maketitle
\begin{abstract}
  We study regret-minimizing online algorithms based on potential
functions. First, we show that any algorithm with a regret bound that
holds for any $\epsilon$ is equivalent to a potential minimizing
algorithm and vice versa. Second we should a min-max learning
algorithm for known horizon. We show a regret bound that is close to
optimal when the horizon is not known. Finally we give an algorithm
with second order bounds that characterize easy sequences.
\end{abstract}

\section{Introduction}
Online prediction with expert advise has been studied extensively over
the years and the number of publications in the area is vast (see
e.g.~\cite{vovk1990aggregating, feder1992universal,
  littlestone1994weighted, cesa1997use, cesa2006prediction}.

Here we focus on a simple variant of online prediction with expert
advice called {\em the decision-theoretic online learning game}
(DTOL)~\cite{freund1997decision}. DTOL (Figure~\ref{fig:DTOL}) is a
repeated zero sum game between a {\em learner} and an {\em
  adversary}. The adversary controls the losses of $N$ actions, while
the learner controls a distribution over the actions.

\begin{figure}[ht!]
\framebox{
\begin{minipage}[t]{6.4in}
For $i=1,\ldots,T$
\begin{enumerate}
    \item The learner chooses a weight function $\l(i,j)$ over the
      actions $j \in \{1,\ldots,N\}$ \\
      such that $\sum_{j=1}^N \l(i,j)=1$  
    \item The adversary chooses an {\em instantaneous loss} for each
      of the $N$ actions: \\
      $l_j^i \in [-1,+1]$ for $j \in \{1,\ldots,N\}$.
    \item The {\em cumulative loss of action $j$} 
    is  $L^i_j = \sum_{s=1}^i l_j^s$. 
    \item The learner incurs an {\em instantaneous average loss} defined as
      $\ell^i = \sum_{j=1}^N \l(i,j) l_j^i$
    \item The {\em cumulative loss of the learner} is
      $L_\ell^i = \sum_{s=1}^i \ell^s$
    \item The {\em cumulative regret} of the learner with respect to
      action $j$ is
      $\R_j^i = L_\ell^i -L_j^i $.
\end{enumerate}
\end{minipage}}
\caption{Decision theoretic online learning \label{fig:DTOL}}
\end{figure}

The goal of the learner (in the percentile version of the game) is to
perform almost as well as $k$ best actions. Specifically, we sort the
regrets in decreasing order
$\R_1^i \geq \R_2^i \geq \cdots \geq \R_k^i \geq \cdots$ and define
$\R_k^i$ to as the regret relative to the $\epsilon=k/M$ top
percentile, denote $\R_\epsilon^i$. Our goal is to find algorithms
that guarantee small upper bounds on $\R_\epsilon^T$. Known bounds
have the form $c \sqrt{T \ln {1/\epsilon}}$, but the algorithm has to
be tuned based on prior knowledge of $\epsilon$. We seek algorithms
with regret bounds that hold {\em simultaneously} for all values of
$\epsilon$. In other words algorithms that do not need to know
$\epsilon$ ahead of time. The following definition formalizes
the concept of simultaneous bounds:
\begin{definition}[Simultaneous regret
  bound (SRB)] \label{def:unif-regret-bound} Let $G: \reals \to [0,1]$ be a
  non-increasing function which maps regret bounds to probabilities.
  A distribution over regrets $\state$ is simultaneously bound by $G$ if
  \[
    \forall \R \in \reals \;\; \P{\rho \sim \state}{\rho \geq \R} \leq G(\R)
  \]
\end{definition}

A potential function is an increasing function
$\pot:\reals \to \reals$. Online learning algorithms based on
potential functions control the regret by upper bounding the average
potential $\E{\R \sim \state}{\pot(\R)}$. 

\begin{definition}[Average potential bound (APB)] \label{def:aver-potential-bound}
  A distribution over he reals $\state$ satisfies the average
  potential function $\pot$ if
  $$\score \doteq \E{\R \sim \state}{\pot(\R)} \leq 1$$
  Where $\pot: \reals \to \reals^+$ is a non decreasing function. The score $\score$ is as defined.
\end{definition}

Potential functions have long been used to design and analyze online
learning algorithms. However, the choice of the potential function has
been somewhat ad hoc.  Here we show a strong one to one relationship
between SBR functions and potential functions:
\begin{theorem}\label{thm:simulBoundAveragePot}
 A distribution $\state$ is simultaneously bounded by $B$ if and only
 if it satisfies the average potential bound with $\phi(\R) = B(\R)^{-1}$
\end{theorem}
The proof of the theorem is in Appendix~\ref{proof:simulBoundAveragePot}.

Theorem~\ref{thm:simulBoundAveragePot} justifies our focus on
potential functions and our quest to find the ``best'' potential
function. 

We start with the bounded horizon case. Fixing a potential function at
the end of the game $\pot(T,\R)$ and the strategies used by the
learner and the adversary, we define potential functions $\pot(i,\R)$
for iterations $i = T-1,T-2,\ldots,0$ such that the score $\score(t)$ is guaranteed to 
be equal for all of the iterations.
\[
  \score(T) = \score(T-1)=\cdots=\score(0)
\]
This allows us to analyze the game one iteration at a time and
construct good strategies for both sides. We name this potential based
game the {\em Integer Time Game}, the analysis of this game is given
in Section~\ref{sec:int-time-game}. The analysis assumes only that the
final potential $\pot(T,\R)$ is strictly positive and has strictly positive
first and second derivatives (We denote the set
of functions that have $0,\ldots,k$ strictly positive derivatives by $\SP{k}$,
the formal definition is given in Section~\ref{sec:preliminaries})

The strategies yielded by the analysis guarantee bounds on the final
score. The adversarial strategy guarantees
$\lowerscore{} \leq \score(T)$, while the learner's strategy
guarantees $\score(T) \leq \upperscore{}$. Unfortunately, these bounds
don't match, i.e. $\lowerscore{}<\upperscore{}$. In other words our
proposed strategies are not min-max optimal. The question of whether
there exist min/max strategies for the integer time game is open.

To find min/max strategies we expand the game. We call the expanded
game the {\em discrete time game}. The expansion involves giving the
more options to the adversary, but not to the learner. As a result,
any upper bound $\upperscore{}$ that holds for a learner strategy in
the discrete time game also holds in the integer time game.

The added option for the adversary is to declare, at the beginning of
each iteration, the range of values of the instantanous losses. In the
integer time game this range is set to $[-1,+1]$. In the discrete time
game the range is chosen by the adversary on iteration $i$ to be
$[-s_i,+s_i]$ for $1\geq s_i >0$. To keep the game balanced between
the adversary and the learner we replace the iteration number $i$ with
real valued {\em time} parameter and let $t_{i+1}=t_i+s_i^2$. this and
another necessary adjustment are described in
Section~\ref{sec:discrete}. Section~\ref{sec:disc-game-strategies}
describes strategies used for the discrete game which are scaled
versions of the strategies for the integer time game.

We fix the potential at the end of the game $\finalPotR \in \SP{4}$
and consider a sequence of adversarial and learner strategies indexed
by $k$: $\adversMdk,\learnerMdk$, where
$\forall i,\; s^k_i = \frac{\sqrt{\realT}}{2^k}$ for some constant
$\realT$. We prove two facts regarding the limit $k \to \infty$.  The
first (Thm.~\ref{thm:seq-of-adv-strategies}) is that
$\lim_{k \to \infty}\upperpotMdk - \lowerpotMdk \to 0$.  The second
(Thm.~\ref{thm:smallerSteps}) is that, if $\pot(T,\R)$
$$\forall k, \forall 0\leq i \leq 2^{2k},t_i=i 2^{-2k} \realT,\; \forall\R,\;\;\;
,\lowerpotMdkpar{k+1}\left(t_i,R\right) >
\lowerpotMdk\left(t_i,R\right)$$ Taken together these facts imply
that, if the fixed potential function for the end of the game
$\finalPotR$ is in $\SP{4}$, then there exists a potential function
$\pot(t,\R)$. The adversarial strategy corresponding to this potential
function corresponds to Brownian motion.  The backwards recursion used
to computer the potential for $t \leq \realT$ is a partial
differential equation known as the Kolmogorov Backward equation.

The main result of this paper is that a {\em single} adversarial
strategy, i.e. Brownian motion, is optimal for any
sufficiently convex potential functions.

The discrete time game presents the adversary with a dilemma.  On the
one hand, The adversary has to declare, on each iteration, an upper bound on the
range of the losses $[-s_i,s_i]$ where $s_i>0$. On the other hand, it
wants to set $s_i$ as small as possible.~\footnote{The situation is
  similar to a folklore game in which each player writes down a number
  on a piece of paper and the player with the largest number wins.}

We introduce a variant of the game called the {\em continuous time
  game} to alleviate this dilemma. In this came the adversary does not
announce the step size and the learner behaves as if
the step size is infinitesimally small. In this case time is advanced according
to the variance of the actual losses. This much more natural algorithm yields
a regret bound that depends on the cumulative variance and is smaller
for easy, low variance sequences.

Until this point our theory holds for any final potential function in $\SP{4}$. We conclude by analyzing two specific potentials.
\begin{enumerate}
\item We derive a potential function and a corresponding learning algorithm
  that is min/max optimal for a given time horizon $\realT$.
  The optimality is in the sense that the
  simultanous regret bound for time $\realT$ has a matching
  simultanous lower bound.
\item By finding solutions to the Kolmogorov Backward equation that
  hold for all $t>0$ we eliminate the need to define a final
  potential. As a result we get an ``anytime'' learning algorithm that
  can be stopped at any time. The specific potential we analyze is
  Normal-Hedge~\cite{}. NormalHedge is not min/max otimal for any
  time, but it is almost optimal for all times.
\end{enumerate}

\section{related work}
Most of the papers on potential based online algorithms consider
one or a few potential functions. Most common is the exponential
potential, but others have been considered~\cite{cesa2006prediction}.
A natural question is what is the difference between potential
functions and whether some potential function is ``best''.

In this paper we consider a large set of potential functions,
specifically, potential functions that are strictly positive and have
strictly positive derivatives of orders up to four. The exponential
potential and the NormalHedge potential~\cite{chaudhuri2009parameter,luo2015achieving}
are member of this set. 

To analyze these potential functions we define a different
game, which we call the ``potential game''. In this game the primary
goal of the learner is not to minimize regret, rather, it is to
minimize the final score $\score^T$. To do so
we define potential functions for intermediate steps: $0 \leq t
<T$.\footnote{The analysis described here builds on a long line of
  work. Including the Binomial Weights algorithm and it's
  variants~\cite{cesa1996line,abernethy2006continuous,abernethy2008optimal}
  as well as drifting games~\cite{schapire2001drifting,freund2002drifting}.}

Zero-order bounds on the regret ~\cite{freund1999adaptive} depend only on $N$
and $T$ and typically have the form
\begin{equation} \label{eqn:0-order-bound}
  \max_j \R_j^T < C E \sqrt{T \ln N}
\end{equation}
for some small constant $C$ (typically smaller than 2).
These bounds can be extended to infinite sets of actions by defining
the $\epsilon$-regret of the algorithm as the regret with respect to
the best (smallest-loss) $\epsilon$-percentile of the set of actions.

this replaces the bound~(\ref{eqn:0-order-bound}) with 
\begin{equation} \label{eqn:0-epsilon-order-bound}
  \max_j \R_j^T < C E \sqrt{T \ln \frac{1}{\epsilon}}
\end{equation}

Lower bounds have been proven that match these upper bounds up to a
constant. These lower bounds typically rely on constructions in which
the losses $l_j^i$ are chosen independently at random to be either
$+1$ or $-1$ with equal probabilities.

Several algorithms with refined upper bounds on the regret have been
studied. Of those, the most relevant to our work is a paper by 
Cesa-Bianchi, Mansour and
Stoltz~\cite{cesa2007improved} on second-order regret bounds.
The bound given in Theorem~5 of ~\cite{cesa2007improved} can be
written, in our notation, as:
\begin{equation} \label{eqn:2nd-order-bound}
  \max_j \R_j^T \leq 4\sqrt{V_T \ln N} +2 \ln N +1/2 
\end{equation}
Where
\[
  \var_i = \sum_{j=1}^N P^i_j (l_j^i)^2 -  \left( \sum_{j=1}^N P^i_j
    l_j^i \right)^2 \mbox{ and } \V_T= \sum_{i=1}^T \var_i
\]

A few things are worth noting. First, as $|l_j^i|\leq 1$,
$\var_j\leq 1$ and therefor $V_T\leq T$. However $\V_T/T$ can be
arbitrarily small, in which case inequality~\ref{eqn:2nd-order-bound}
provides a tighter bound than ~\ref{eqn:0-order-bound}. Intuitively,
we can say that $\V_T$ replaces $T$ in the regret bound. This paper
provides additional support for replacing $T$ with $\V_T$ and provides
lower and upper bounds on the regret involving $\V_T$.

\section{Main Results}
\begin{enumerate}
\item {\bf Uniform regret bound} There exists an online learning
  algorithm such that for any $\nu>0$ (set in advance) and any
  $t,\epsilon$ (holds uniformly) the following regret bound holds.
  \begin{equation}
\R_\epsilon \leq \sqrt{(t+\nu) \left( \ln (t+\nu) + 2 \ln \frac{1}{\epsilon}\right)}
\end{equation} 
\item {\bf Second order bound}
\item {\bf optimality of Brownian motion} For any potential function
  in $\SP{4}$ the min/max value of any state $(t,R)$ is attained by
  Brownian motion on the part of the adversary for any $s\geq t$.
\end{enumerate}

\section{Preliminaries} \label{sec:preliminaries}

We define some terms and notationthat will be used in the rest of the paper.

{\bf Positivity}
We require that potential functions have positive derivatives for a
range of degree. To that end we use the following definition:
\begin{definition}[Strict Positivity of degree $k$]
A function $f:\reals \to \reals$ is strictly positive of degree $k$, 
denoted $f \in \SP{k}$ if the derivatives of orders 0 to $k$:  
$f(x), \frac{d}{dx}f(x), \ldots, \frac{d^k}{dx^k}f(x)$ exist and are strictly positive.
\end{definition}
The following useful lemma states that $\SP{k}$ is closed under positive conbinations.
\begin{lemma}  \label{lemma:SP-pos-comb}
  Suppose that for $i =1,\ldots,n$, $f_i \in \SP{k}$ and $\alpha_i>0$,
  Then $\sum_{i=1}^k \alpha_i f_i \in \SP{k}$
\end{lemma}

{\bf Divisibility:} To reach optimality we need the set of actions to
be arbitrarily divisible. Intuitively, We replace the finite set of
actions with a continuous mass, so that each set of actions can be
partitioned into two parts of equal weight.  Formally, we define the
set of actions to be a probability space $(\Omega,\sigma,\mu)$ such
that $\omega \in \Omega$ is a particular action. We require that the
space is {\em arbitrarily divisible}, which means that for any
$s \in \sigma$ , there exist a partition $u,v \in \sigma$ such that
$u \cup v = s, u \cap v = \emptyset$, and
$\P{}{u}=\P{}{v}=\frac{1}{2} \P{}{s}$.

{\bf State:} The {\em state} of a game at iteration $i$, denoted $\state(i)$, is
a random variable that maps each action $\omega \in \Omega$ to the
cumulative regret of $\omega$ at time $i$: $\R_\omega^i$. The sequence
of cumulative regrets corresponding to action $\omega$ is the {\em
  path} of $\omega$:
\begin{equation} \label{eqn:path}
  S_{\omega}=(\R_\omega^1,\R_\omega^2,\ldots,\R_\omega^N)
\end{equation}

{\bf  Generalized binomial distribution}
We denote by $\Binom(n,s)$ the distribution over the reals defined by
$\sum_{i=1}^n X_i$ where $X_i$ are iid binary random variables which
attain the values $-s,+s$ with equal probabilities.

{\bf Expected value shorthand:} Suppose $P$ is a distribution over the reals, and $f:\reals
\to \reals$, we use the following short-hand notation for the expected
value of $f$ under the distribution $P$:
\[ P \odot f \doteq \E{x \sim P}{f(x)}  \]
We define the {\em score} at iteration $i$ as the average potential
with respect to the state:
\[ \score(i) = \state(i) \odot \pot(i) \doteq \E{\R \sim \state(i)}{\pot(i,\R)}\]
Note that in this short-hand notation we suppress the variable with
respect to which the integration is defined, which will always be $\R$.

{\bf Convolution:} Let $A,B$ be two independent random variables. We define the
convolution $A \oplus B$ to be the distribution of $x+y$. A constant
$a$ corresponds to the point mass distribution concentrated at
$a$. For convenience we define $A \ominus B = A \oplus (-B)$

\section{Integer time game}
\label{sec:int-time-game}
The integer time game is described in
Figure~\ref{fig:integerTimeGame}.  The integer time game generalizes
the decision theoretic online learning problem~\cite{FreundSc97} in
the following ways:
\begin{enumerate}
\item The goal of the learner in DTOL is to guarantee an upper bounds
  on the regret. The learner's goal in the integer time game is to
  minimize the final score. From theorem~\ref{thm:simulBoundAveragePot} we know that if
  we set the final potential as $\finalPotT(\R) = \frac{1}{G(\R)}$ then the two
  conditions are equivalent, allowing us to focus on the score.
\item The number of iterations $T$ is given as input, as is the
  potential function at the end: $\finalPotT(\R)$.
\item The actions are assumed to be {\em divisible}. For our purposes
it is enough to assume that any action can be split into two equal
weight parts.
\end{enumerate}

The key to the potential based analysis is that using the predefined
final potential we can define potential functions and scores for all
iterations $1,\ldots,T-1$. This is explained in the next subsection.

\begin{figure}[ht]
\framebox{
\begin{minipage}[t]{6.4in}
Initialization:
\begin{itemize}
\item input: $T$ : The number of iterations.
\item Final iteration potential function:  $\finalPotT \in \SP{2}$
\item $\state(1) = \delta(0)$ is the initial state of the game which
  is a point mass distribution at 0. 
\end{itemize}

For $i=1,2,\ldots,T$:\\

\begin{enumerate}
\item The learner chooses a non-negative random variable over $\Omega$
  that is the {\em weight function} $\learnerM(i,\R)$ such that
  $\state(i) \odot \learnerM(i)=1$
\item The adversary chooses a function $\adversM(i,\R)$ that maps
  $i,\R$ to a distribution over $[-1,+1]$. This
  random variable corresponds to the instantaneous loss of each action at
  time $t$.
\item 
  We define the {\em bias} at $(i,\R)$ to be
  \begin{equation} \label{eqn:Bias}
    \Bias(i,\R) \doteq \E{l \sim \adversM(i,\R)}{l}
  \end{equation}
\item the average loss is 
  \begin{equation} \label{eqn:aggregate-loss}
    \ell(i)=\state(i,\R) \odot \paren{\learnerM(i,\R) \Bias(i,\R)}
  \end{equation}
\item The state is updated. 
  \begin{equation} \label{eqn:state-update}
    \state(i+1) = \E{\R \sim \state(i)}{\R \oplus  \adversM(i,\R)}
    \oplus -\ell(i)
    \end{equation}
  Where $\adversM(i,\R)$ is the distribution of the losses of actions
  with respect to which the regret is $\R$ after iteration
  $i-1$. $\oplus$ denotes the convolution as defined above.
\end{enumerate}

The final score is calculated: $\score(T)=\state(T) \odot
\finalPotT$.

The goal of the learner is to minimize this score, the goal
of the adversary is to maximize it.
\end{minipage}}
\caption{The integer time game \label{fig:integerTimeGame}}
\end{figure}

\subsection{Defining potential Functions for all iterations\label{sec:potentials}}

The potential game defines the {\em final} potential function
$\finalPotT$, at the end of the game. We will now show, that we can
extend the definition of a potential function to all iterations of the game.

 A single action defines a path $S_\omega$ (as defined in (\ref{eqn:path})). Fixing
 the strategies of the learner and the adversary determines
 a distribution $\D$ over paths.
We describe two equivalent ways to define $\potPQ(i,\R)$ for $i<T$
 \begin{enumerate}
 \item{\bf Using conditional expectation} We can define the potential
   on iteration $i$ based on the fixed potential at iteration $T$.
\begin{equation}
  \forall i=1,\ldots,T,\; \forall \R \;\;\;
  \potPQ(i,\R)=\E{\omega\sim \D|\R_\omega^i=\R}{\pot(T,\R_{\omega}^T)}
\end{equation}
 \item{\bf Using backward induction} It is sometimes convenient to
   compute the the potential for time $i$ from the potential at time
   $i+1$:
   \begin{equation} \label{eqn:back-induction}
     \forall i=1,\ldots,T-1, \R\;\;\;
     \potPQ(i,\R)=\E{\omega\sim \D|M_\omega^i=\R}{\potPQ(i+1,\R_{\omega}^{i+1})}
   \end{equation}
by using backwards induction: $i=T-1,T-2,\ldots,1$ we can compute the
potential for all iterations.

We use Equations~(\ref{eqn:Bias},\ref{eqn:aggregate-loss}) and
marginalizing over $\R$ to express Equation~(\ref{eqn:back-induction})
in terms of the single step strategies:
\begin{equation} \label{eqn:induction}
  \forall i=1,\ldots,T-1, \R\;\;\;
  \potPQ(i,\R) \doteq \E{r \sim [(\R-\ell(i)) \oplus \adversM(i,\R)]}{\potPQ(i+1,r)}
\end{equation}

\end{enumerate}

The score at iteration $i$ is defined as
$\score(i)=\state(i)\odot \pot(i)$. The scores are all different
expressions for calculating the expected final potential for the fixed strategies
$\adversM, \learnerM$. Therefor the scores are all equal, as expressed
in the following theorem:

\begin{theorem} \label{thm:backward-recursion}
Assuming $ \learnerM(i,\R), \adversM(i,\R)$ are fixed for all
$i=1,\ldots,T-1$, then
\[
  \state(T)\odot \pot(T)=\score(T)=\score(T-1)=\cdots=\score(1)=\potPQ(0,0)
  \]
\end{theorem}

A few things worth noting:
\begin{enumerate}
\item $\potPQ(i,\R)$ is the the final expected potential
  given that the paths starts at $(i,\R)$ and that
  the strategies used by both players in iterations $i,\ldots,T$ are fixed. Note
  also that which strategies were used in iterations $1,\ldots,i-1$ is
  of no consequence. The effect of past choices is captured by the
  state $\state(i)$.
\item
  The final expected potential is equal to $\pot(0,0)$ which is the
  potential at the common starting point: $i=1$, $\R=0$.
\end{enumerate}

\subsection{Upper and Lower potentials}
Next,we vary the strategies of one side or the other to define upper
and lower potentials.
\begin{equation} \label{eqn:upperPotentials}
  \exists \learnerM, \;\;\; \forall \adversM,\;\; \forall 1\leq i \leq
  T,\;\; \forall \R \in \reals,\;\; \upperpot(i,\R) \geq \potPQ(i,R)
\end{equation}
\begin{equation} \label{eqn:lowerPotentials}
  \exists \adversM, \;\;\; \forall \learnerM, \;\; \;\; \forall 1\leq i \leq
  T,\;\; \forall \R \in \reals,\;\; \lowerpot(i,\R) \leq \potPQ(i,R)
\end{equation}

In words, $\upperpot$ is an upper bound on the potential that is 
guaranteed by the learner strategy $\learnerM$ while $\lowerpotd$
is a lower bound that is guaranteed by the adversarial
strategy $\adversM$.

Following the same argument as the one leading to
Theorem~\ref{thm:backward-recursion}. We define upper and lower scores
$\upperscoreM(i),\lowerscoreM(i)$ such that
\begin{equation} \label{eqn:upper-recursion}
  \state_{\learnerM}(T)\odot \finalPotT=\upperscoreM(T)=\upperscoreM(T-1)=\cdots=\upperscoreM(0)=\upperpot(0,0)
\end{equation}
and
\begin{equation} \label{eqn:lower-recursion}
  \state_{\adversM}(T)\odot \finalPotT=\lowerscoreM(T)=\lowerscoreM(T-1)=\cdots=\lowerscoreM(0)=\lowerpot(0,0)
\end{equation}

Our ultimate goal is to find strategies $\learnerM$ and
$\adversM$ such that
\begin{equation} \label{eqn:limitPotential}
\forall i,\R,\;\;\; \lowerpot(i,\R) = \upperpot(i,\R)
\end{equation}
in particular, $\lowerscoreM(0)=\lowerpot(0,0) =
\upperpot(0,0)=\upperscoreM(0)$. This means that
$\adversM,\learnerM$ are a min/max pair of strategies and that
$\lowerscoreM(0)=\upperscoreM(0)$ define the min/max value of the game.
~\\~\\
We do not achieve this for the integer game described in the next
section. To achieve min/max optimality we extend the integer time game
to the discrete time game (section~\ref{sec:discrete}) and to the
continuous time game (\ref{sec:continuous}).

\subsection{Strategies for the integer time  game} \label{sec:strat-integer}
We assume that $\finalPotT  \in \SP{2}$, in other words, the final
potential is positive, increasing and convex. $\finalPotT$ defines the
upper and lower potentials at time $T$:
$$\lowerpotb(T,\R) = \upperpotb(T,\R) = \finalPotT(\R) $$
We define a backwards recursion for the lower potential:
\begin{equation} \label{eqn:backward-iteration-lower}
  \lowerpotb(i-1, \R) = \frac{\lowerpotb(i,\R+1) + \lowerpotb(i,\R-1)}{2}
\end{equation}
and a backwards recursion for the upper potential:
\begin{equation} \label{eqn:backward-iteration-upper-recursion}
  \upperpotb(i-1, \R) = \frac{\upperpotb(i,\R+2) + \upperpotb(i,\R-2)}{2}
\end{equation}

We define strategies that correspond to these potentials. A strategy
for the adversary:
\begin{equation} \label{eqn:adv-strat-p}
  \adversMb(i,\R) =
  \begin{cases}
    +1 & \mbox{ w.p. } \frac{1}{2}\\
    -1 & \mbox{ w.p. } \frac{1}{2}\\
  \end{cases}
\end{equation}
and a strategy for the learner:
\begin{equation} \label{eqn:learner-strat-1}
\learnerMb(i,\R) = \frac{1}{Z} \frac{\pot(i,\R+2) - \pot(i,\R-2)}{2}
\end{equation}
Where $Z$ is a normalization factor
$$Z = \E{\R \sim \state(i)}{\frac{\pot(i,\R+2) - \pot(i,\R-2)}{2}}$$

The following lemma states that these strategies guarantee the
corresponding potentials.
\begin{lemma} \label{lemma:first-order-bound}
~\\
Let $i$ be an integer between $1$ and $T$

If $\lowerpotb(i,\R) \in \SP{2}$
\begin{enumerate}
\item {\bf Positivity:} $\lowerpotb(i-1,\R) \in \SP{2}$
\item {\bf Adversary:} The adversarial strategy~(\ref{eqn:adv-strat-p})
  guarantees the recursion given in Eq.~(\ref{eqn:backward-iteration-lower})
\end{enumerate}

If $\upperpotb(i,\R) \in \SP{2}$
\begin{enumerate}
\item {\bf Positivity:} $\upperpotb(i-1,\R) \in \SP{2}$
\item {\bf Learner:} The learner strategy~(\ref{eqn:learner-strat-1})
  guarantees the recursion given in Eq.~(\ref{eqn:backward-iteration-upper-recursion})
\end{enumerate}

\end{lemma}

\proof We prove each claim in turn
\begin{enumerate}
\item {\bf Positivity:} Follows from Lemma~\ref{lemma:SP-pos-comb}.
\item{\bf Adversary:} By symmetry adversarial strategy~(\ref{eqn:adv-strat-p}) guarantees that
  the aggregate loss~(\ref{eqn:aggregate-loss}) is zero regardless of
  the choice of the learner: $\ell(i)=0$.
  Therefor the state update~(\ref{eqn:state-update}) is equivalent to
  the symmetric random walk:
  $$\state(i) = \frac{1}{2} \paren{(\state(i) \oplus 1) + (\state(i)
    \ominus 1)}$$
  Which in turn implies that if the adversary plays $\adversM^*$
  and the learner plays an arbitrary strategy $\learnerM$
  \begin{equation} \label{eqn:lower}
    \lowerpotb(i-1,\R) = \frac{\lowerpotb(i,\R-1)+\lowerpotb(i,\R+1)}{2}
  \end{equation}
  As this adversarial strategy is oblivious to the learner's strategy, it
  guarantees that the average value at iteration $i$ is {\em equal} to the
  average of the lower value at iteration $i$.
\item {\bf Learner:}
  Plugging learner's strategy~(\ref{eqn:learner-strat-1})
  into equation~(\ref{eqn:aggregate-loss}) we find that
 \begin{equation} \label{eqn:ell-optimal-learner}
   \ell(i) = \frac{1}{Z_{i}} \E{\R \sim \state(i)}{\paren{\upperpotb(i,\R+2)-\upperpotb(i,\R-2)}
   \Bias(i,\R)}
\end{equation}
  Consider the score at iteration $i$ when the learner's strategy
  is $\learnerM^*$ and the adversarial strategy  $\adversM$ is arbitrary
     \begin{equation} \label{eqn:Pot-Update}
    \score_{\learnerM^*,\adversM}(i,\R) = \E{\R \sim \state(i)}{ \E{y \sim
      \adversM(i)(\R)}{\pot(i,\R+y-\ell(i))}}
  \end{equation}
  As $\pot(i,\cdot)$ is convex and as $y-\ell(i) \in [-2,2]$,
  \begin{equation} \label{eqn:pot-upper}
    \upperpotb(i-1,\R+y) \leq \frac{\upperpotb(i,\R+2)+\upperpotb(i,\R-2)}{2} +
    (y-\ell(i)) \frac{\upperpotb(i,\R+2)-\upperpotb(i,\R-2)}{2}
    \end{equation}
  Combining the equations~(\ref{eqn:ell-optimal-learner}) and~(\ref{eqn:Pot-Update}) we find that
  \begin{eqnarray}
  \score_{\learnerM^*,\adversM}(i,\R)&=&\E{\R \sim \state(i)}{\E{y \sim \adversM(i)(\R)}{\upperpotb(i,\R+y-\ell(i))}}\\
  &\leq & \E{\R \sim \state(i)}{\frac{\upperpotb(i,\R+2)+\upperpotb(i,\R-2)}{2}}\\
  &+&
  \E{\R \sim \state(i)}{\E{y \sim \adversM(i)(\R)}{(y-\ell(i)) \frac{\upperpotb(i,\R+2)-\upperpotb(i,\R-2)}{2}}} \label{eqn:zero-term}
  \end{eqnarray}
  
The final step is to show that the term~(\ref{eqn:zero-term}) is equal
to zero. As $\ell(i)$ is a constant with respect to $\R$ and $y$ the
term~(\ref{eqn:zero-term}) can be written as:
\begin{eqnarray}
&&\E{\R \sim \state(i)}{\E{y \sim \adversM(i)(\R)}{(y-\ell(i))
   \frac{\upperpotb(i,\R+2)-\upperpotb(i,\R-2)}{2}}}\\
&=&
\E{\R \sim \state(i)}{\Bias(i,\R)
    \frac{\upperpotb(i,\R+2)-\upperpotb(i,\R-2)}{2}}\\
  &-& \ell(i) \E{\R \sim \state(i)}{
    \frac{\upperpotb(i,\R+2)-\upperpotb(i,\R-2)}{2}}\\
  &=& 0
\end{eqnarray}
\end{enumerate}
\qed

Repeating the induction steps of Lemma~\ref{lemma:first-order-bound}
from $i=T$ to $i=1$ yields the following theorem.
\begin{theorem} \label{thm:IntegerGameBounds}
  Let $\finalPotT \in \SP{2}$, for any iteration $0 \leq i \leq T$ and
  regret $\R_0 \in \reals$ 
  \begin{itemize}
  \item
    The lower potential guaranteed by $\adversMb$ is
     $$\lowerpotb(i,R_0) = \E{\R \sim \R_0 \oplus \Binom(T-i,1)}{\finalPotT(\R)} $$
  \item
    The upper potential guaranteed by $\learnerMb$ is
    $$\upperpotb(i,R_0) = \E{\R \sim \R_0 \oplus \Binom(T-i,2)}{\finalPotT(\R)}$$
  \end{itemize}
\end{theorem}

Plugging in $i=0$,$\R=0$ we get the following Corrolary:
\begin{corollary}
  if the learner plays $\learnerMb$ on every iteration
  it guarantees that the final score satisfies
  $$\state(T) \odot \finalPotT \leq \Binom(T,2) \odot \finalPotT $$

  If the Adversary plays $\adversMb$ on every iteration it guarantees
  that:
  $$\state(T) \odot \finalPotT = \Binom(T,1) \odot \finalPotT $$
\end{corollary}
\section{From integer to discrete time}
\label{sec:discrete}

The upper and lower bound on the final score given in
Theorem~\ref{thm:IntegerGameBounds} do not match. If
$\finalPotT \in \SP{2}$ then \linebreak
$\Binom(T,1) \odot \finalPotT <\Binom(T,2) \odot \finalPotT$ In other
words, the
strategies~(\ref{eqn:adv-strat-p},\ref{eqn:learner-strat-1}) are not a
min/max pair.\footnote{There might be other (pure) strategies for the
  integer game that are a min/max pair, we conjecture that is not the
  case, and seek a extension of the game that would yield min/max
  strategies.}

To close this gap we extend the integer time game into a new game we
call the discrete time game (Fig.~\ref{fig:discrete-Time-Game}). The
discrete time game increases the options available to the adversary,
but not to the learner.  As the integer step game is a special case of
the new game, any upper potential that can be guaranteed by the
learner in the discrete time game is also an upper potential for the
discrete time game.

In the integer time game the loss of each action is in the range
$[-1,+1]$, in the discrete time game the adversary chooses, on
iteration $i$ a step size $0<s_i\leq 1$ which restricts the losses to
the range $[-s_i,+s_i]$. Note that by always choosing $s_i=1$,
the adversary can choose to play the integer time game.

\begin{figure}[ht!]
\framebox{
\begin{minipage}[t]{6.4in}
Initialization: $t_0=0$ \newline
  
On iteration $i=1,2,\ldots$
\begin{enumerate}
\item  If $t_{i}= \realT$ the game terminates.
\item The adversary chooses a {\em step size} $0<s_i\leq \min(\sqrt{1-t_i},1)$, which advances
  time by $t_i = t_{i-1}+s_i^2$
\item Given $s_i$, the learner chooses a distribution $\learnerM(i)$ over $\reals$.
\item The adversary chooses a mapping from $\reals$ to distributions
  over $[-s_i,+s_i]:\;\adversM(t,\cdot): \reals \to \Delta^{[-s_i,+s_i]}$
  
\item The aggregate loss is calculated:
  \begin{equation} \label{eqn:ell-discrete}
    \ell(t_i)=\E{\R \sim \state(t_i)}{\learnerM(t_i,\R) \Bias(t_i,\R)}
    \mbox{ where } \Bias(t_i,\R) \doteq \E{y \sim \adversM(t_i,\R)}{y}
  \end{equation}
  Such that $|\ell(t_i)| \leq s_i^2$
\item The state is updated. 
  $$\state(t_i) = \E{\R \sim \state(t_{i})}{\adversM(t_i,\R) \oplus (\R-\ell(t_i))}
  $$
  Where $\oplus$ is a convolution as defined in the preliminaries.
\end{enumerate}

Upon termination, the final value is calculated:
$$\score(\realT) =\state(\realT) \odot \pot(\realT)$$

\end{minipage}}
\caption{The discrete time game  \label{fig:discrete-Time-Game}}
\end{figure}

We make two additional alterations to the integer time game in order to
keep the game fair. An unfair game is one where one side always wins.
We list the alterations and then justify them.
\begin{enumerate}
\item {\bf real-valued time} In the integer time game we use an
  integer to indicate the iteration number: $i=1,2,\ldots,T$. In the
  discrete time game we use an positive real value, which we call
  ``time'' and use the update rule $t_{i+1} = t_i + s_i^2$, and define
  the final time, which is used in the regret bound, to be
  $\realT=\sum_{i=0}^T s_i^2$
\item {\bf Bounded average loss} We restrict the average loss to a
  range much smaller than $[-s_i,+s_i]$, specifically:
  $|\ell(i)| \leq s_i^2$
\end{enumerate}
Note that both of these conditions hold trivially when $s_i=1$
\begin{enumerate}
  \item {\bf Justification of real-valued time}
To justify these choices we consider the following adversarial
strategy for the discrete time game:
\begin{equation} \label{eqn:adv-strat-s}
  \adversMd \brac{s,p} \paren{t,R} =
  \begin{cases}
    +s & \mbox{ w.p. } p\\
    -s & \mbox{ w.p. } 1-p\\
  \end{cases}
\end{equation}

From Equation~(\ref{eqn:lower-recursion}) we get that the initial
score,
\[
  \lowerscoreMd(0) = \lowerscoreMd(T)=\state_{\adversMd}(T) \odot \pot(T)
\]

On the other hand, we know that  $\state_{\adversMd}(T) =
\Binom(T,s)$. Suppose $T$ is large enough that the normal approximation for the
binomial can be used. Let ${\cal N}(\mu,\sigma^2)$ be the normal
distribution with mean $\mu$ and variance $\sigma^2$.
\begin{equation}   \label{eqn:largeTlimit}
  \lim_{T \to \infty} \lowerscoreMd(0) =  {\cal N}(0,Ts^2) \odot \pot(T)
\end{equation}

Recall that $\pot(T)$ is a fixed strictly convex function. It is not
hard to see that if $Ts^2 \to 0$ minimizes  $\lowerscoreMd(0)$ and
makes it equal to to $\pot(T,0)$, which means that the learner wins,
while if $Ts^2 \to \infty$, $\lowerscoreMd(0) \to \infty$ which means
that the adversary wins. In order to keep the game balanced
keep $Ts^2$ constant as we let $s \to 0$. We achieve that by defining
the real-valued discrete time as $t_j = \sum_{i=0}^{j-1} s_i^2$.

\item {\bf Justification of bounding average loss} Suppose the game is
  played for $T$ iterations and that the adversary uses the strategy
  $\adversMd\brac{s,\frac{1}{2}+\epsilon}\paren{t,\R}$ and that
  $s=\frac{1}{\sqrt{T}}$. In this case the loss of the learner in
  iteration $i$ is $\ell(i)=2s\epsilon$ and the total loss is
    $$L_\ell^T=\sum_{i=0}^{T-1} \ell(i) = T 2 \epsilon s = \frac{2 \epsilon}{s}$$.

    If $\epsilon$ is kept constant as $s \to 0$
    then $\lim_{T \to \infty}L_\ell^T=\infty$, biasing the game towards the adversary. On the other
    hand, if $\epsilon =s^{\alpha}$ for $\alpha<1$ then $L_\ell^T
    \to 0$, biasing the game towards the learner. To keep the game
    balanced we have to set $\epsilon=cs$ for some constant
    $c$. Without loss of generality we set $c=1$.

    Generalizing this to the game where the adversary can choose a
    different $s_i$ in each iteration we get the constraint 
    $|\ell(i)| \leq s_i^2$
\end{enumerate}

\subsection{Strategies for discrete time}
\label{sec:disc-game-strategies}

We fix a real number $\realT$ as the real length of the game.

We define a sequence of adversarial strategies, indexed by $k$, where the step size of $\adversMdk$ is $s_k = 2^{-2k} \sqrt{\realT}$.

We define a sequence of adversarial strategies $\adversMdk$ and
matching learner strategies $\learnerMdk$ for $k=0,1,2,\ldots$. The
adversarial strategies are designed so that the upper and lower
potentials converge to a limit as $k \to \infty$.

We set the time points $t_i =i s_k^2$ for $i=0,1,\ldots,2^{2k}$. We
call the resulting games $k$-discrete and denote them as $D(k)$.

For a given $k$ we define upper and lower potentials for each
$t_i$. This is done by induction starting with the final potential
function  $\finalPotR(\R)=\lowerpotMdk(\realT, \R)=\upperpotMdk(\realT, \R)$ and iterating
backwards for  $i=T,T-1,\ldots,0$, $t_i = i s_k^2$ 
 \begin{equation} \label{eqn:backward-iteration-lower-discrete}
   \lowerpotMdk(t_{i-1}, \R) = \frac{\lowerpotMdk(t_i,\R+s_k) + \lowerpotMdk(t_i,\R-s_k)}{2}
 \end{equation}

 \begin{equation} \label{eqn:backward-iteration-upper-recursion-discrete}
   \upperpotMdk(t_{i-1}, \R) = \frac{\upperpotMdk(t_i,\R+s_k(1+s_k)) + \upperpotMdk(t_i,\R-s_k(1+s_k))}{2}
 \end{equation}

These upper and lower potentials correspond to strategies for the
adversary and the learner.
The adversarial strategy is 
\begin{equation} \label{eqn:adv-strat-dk}
  \adversMdk=  \begin{cases}
    +s_k & \mbox{ w.p. } \frac{1}{2}\\
    -s_k & \mbox{ w.p. } \frac{1}{2}\\
  \end{cases}
\end{equation}

The learner's strategy is:
\begin{eqnarray} \label{eqn:learner-strat-1c}
  \learnerMdk(t_{i},\R) = \frac{1}{Z}
  \frac{\upperpotMdk(t_{i+1},\R+s_k(1+s_k)) -
  \upperpotMdk(t_{i+1},\R-s_k(1+s_k))}{2} \\
  \mbox{ where } Z = \E{\R \sim \state(t_{i+1})}{\frac{\upperpotMdk(t_{i+1},\R+s_k(1+s_k)) -
  \upperpotMdk(t_{i+1},\R-s_k(1+s_k)}{2}} \nonumber
\end{eqnarray}

The potentials and strategies defined above are scaled versions of the
integer time potential recursions defined in
Equations~(\ref{eqn:backward-iteration-lower},\ref{eqn:backward-iteration-upper-recursion})
and the strategies defined in Equations~(\ref{eqn:adv-strat-p},\ref{eqn:learner-strat-1}). Specifically, the games operate on lattices that we will now describe.

The adversarial strategy $\adversMb$ defines the following lattice over $i$ and $R$:
$$\Ilat{T}=\left\{ (i,2j-i) \left| 0 \leq i \leq T, 0 \leq j \leq i\right. \right\}$$

The $k$'th adversarial strategy $\adversMdk$ uses step size $s_k=\sqrt{\realT} 2^{-k}$ and time
increments $s_k^2=\realT 2^{-2k}$. We define the {\em game lattice}
for $k$ as the set of $(t,R)$ pairs that are reached by $\adversMdk$.
$$\Klat{{\realT,k}}=\left\{ (t,\R) \left| t=i s_k^2, 0 \leq i \leq 2^{2k}, \R=(2j-i)s_k, 0 \leq j \leq i\right. \right\}$$
$\Ilat{T}$ is a special case of $\Klat{\realT,k}$ because setting
$\realT=T=2^{2k}$ we get that $s_k=s_k^2=1$ and  $\Klat{\realT,k} = \Ilat{T}$.

It is not hard to show that the lattices get finer with $k$, i.e. if  $j \leq k$,  $\Klat{\realT,j} \subseteq \Klat{\realT,k}$.

The following Lemma parallels Lemma~\ref{lemma:first-order-bound} for the integer time game.
\begin{lemma} \label{lemma:discrete-step-bound}
~\\
Let $i$ be an integer between $1$ and $T$

If $\lowerpotMdk(t_i,\R) \in \SP{2}$
\begin{enumerate}
\item  $\lowerpotMdk(t_{i-1},\R) \in \SP{2}$
\item The adversarial strategy~(\ref{eqn:adv-strat-dk})
  guarantees the recursion given in Eq.~(\ref{eqn:backward-iteration-lower-discrete})
\end{enumerate}

If $\upperpotMdk(t_i,\R) \in \SP{2}$
\begin{enumerate}
\item $\upperpotMdk(t_{i-1},\R) \in \SP{2}$
\item The learner strategy~(\ref{eqn:learner-strat-1c})
  guarantees the recursion given in Eq.~(\ref{eqn:backward-iteration-upper-recursion-discrete})
\end{enumerate}

\end{lemma}
\proof
The statement of the Lemma and the proof are scaled versions of
Lemma~\ref{lemma:first-order-bound} and its proof. The iteration step
is $s_k^2$ instead of $1$ while the loss/gain of an action in a single
step is $[-s_k,s_k]$ instead of $[-1,+1]$.

One change worth noting is at the step from
Equation~(\ref{eqn:Pot-Update}) and Equation~(\ref{eqn:pot-upper}),
where the bound  $y-\ell(i) \in [-2,2]$ is replace by  $y-\ell(i) \in
[-s_k-s_k^2,s_k+s_k^2]$. This follows from the bound $|\ell(i)| \leq
s_k^2$ which is discussed in Section~\ref{sec:discrete}.
\qed

\begin{theorem} \label{thm:DescreteGameExactValues}
  Let $\finalPotR \in \SP{2}$ be the final potential in the discrete
  time game. Fix $k$ and the step size $s_k=\sqrt{\realT} 2^{-k}$, and let $t_i=i s_k^2$ for $i=0,1,\ldots,2^{2k}$ and
  let $\R_0$ be a real value, then 
  \begin{itemize}
  \item
    The lower potential guaranteed by $\adversMdk$ is
    \begin{equation} \label{eqn:lower-potential-exact}
      \lowerpotMdk(t_i,\R_0) = \E{\R \sim \R_0 \oplus
        \Binom\paren{2^{2k}-i,s_k}}{\finalPotR(\R)}
      \end{equation}
  \item
    The upper potential guaranteed by $\learnerMdk$ is
    \begin{equation} \label{eqn:upper-potential-exact}
    \upperpotMdk(t_i,\R_0) =  \E{\R \sim \R_0 \oplus
      \Binom\paren{2^{2k}-i,s_k(1+s_k)}}{\finalPotR(\R)}
    \end{equation}
  \end{itemize}
\end{theorem}

Using Theorem~\ref{thm:DescreteGameExactValues} we can show that, as
$k \to \infty$, the upper lower potential converge to the same limit.

\begin{theorem} \label{thm:seq-of-adv-strategies}
  ~\\
  Fix $\realT$ and assume $\finalPotR \in \SP{2}$. Consider the sequence of upper
  and lower potentials  $\upperpotMdk,\lowerpotMdk$ for
  $k=0,1,2,\ldots$.

  Then for any  $0 < t \leq \realT$ and any $\R_0$:
  \begin{equation} \label{eqn:k-limit}
    \lim_{k \to \infty} \upperpotMdk(t,\R_0) =
    \lim_{k \to \infty}\lowerpotMdk(t,\R_0)=
    {\cal N}(\R_0,\realT-t) \odot \finalPotR
  \end{equation}
\end{theorem}

\proof

We first assume that $(t,\R_0) \in \Klat{j,\realT}$ and that $k\geq j$. We later expand to any $0 < t \leq \realT$ and any $\R_0 \in \reals$.
Consider Equation~\ref{eqn:upper-potential-exact} for $\learnerMdk$ and $\learnerMdj$,
keeping $t$ and $j$ constant and letting $k \to \infty$.

\begin{equation} \label{eqn:upper-potential-exact-j}
  \upperpotMdj(t,\R_0) =  \E{\R \sim \R_0 \oplus
    \Binom\paren{2^{2j}-i_j,s_j(1+s_j)}}{\finalPotR(\R)}
\end{equation}

\begin{equation} \label{eqn:upper-potential-exact-k}
  \upperpotMdk(t,\R_0) =  \E{\R \sim \R_0 \oplus
    \Binom\paren{2^{2k}-i_k,s_k(1+s_k)}}{\finalPotR(\R)}
\end{equation}
We rewrite the binomial factor in
Eq~(\ref{eqn:upper-potential-exact-k})
$$
 \Binom\paren{2^{2k}-i_k,s_k(1+s_k)} =
 \Binom\paren{2^{2(k-j)}\paren{2^{2j}-i_j},2^{j-k} s_j(1+2^{j-k} s_j)}
 $$
As $j$ is constant, $s_j$ is constant and so is $a_j \doteq
2^{2j}-i_j$. Multiplying the number of steps by the variance per step
we get
$$Var(\Binom_k) = 2^{2(k-j)}a_j\paren{2^{j-k} s_j(1+(2^{j-k} s_j)}^2
= a_j s_j^2 (1+(2^{j-k} s_j))^2 
$$

As $s_j,a_j$ are constants we get that $\lim_{k \to \infty}
Var(\Binom_k) = a_j s_j$.  From the central limit theorem we get that
for any $(t,\R_0) \in \Klat{j,\realT}$
\begin{equation} \label{eqn:conergence-for-k}
  \lim_{k \to \infty} \upperpotMdk(t,\R_0) \odot \finalPotR = {\cal
    N}(\R_0,\realT-t) \odot \finalPotR
\end{equation}

Our argument hold for all $(t,\R_0) \in \bigcup_{k=0}^\infty \Klat{k,\realT}$ which is dense in the set $0<t\leq \realT, \R_0 \in \reals$.
On the other hand, $\upperpotMdk(t,\R) \odot \finalPotR$ is continuous in both
$t$ and $R$, therefor Equation~(\ref{eqn:conergence-for-k})  holds for all $t$ and $\R$.

As similar (slightly simpler) argsument holds for the lower potential limit
$\lim_{k \to \infty} \lowerpotMdk(t,\R_0)$

\qed

We have shown that in the limit $s \to 0$ the learner and the
adversary converge to the same potential function. In the next section
we show that this limit is the min/max solution by describing conditions
under which the adversary prefers using ever smaller steps size.

\subsection{The adversary prefers smaller steps} \label{sec:smallsteps}

Theorem~\ref{eqn:conergence-for-k} characterizes the limits of the
upper and lower potentials, as $k \to \infty$ are equal to each other
and to ${\cal N}(\R_0,\realT-t) \odot \finalPotR$ To show that this
limit corresponds to the min/max slolution of the game we need to show
that the adversary perfers smaller steps. In other words, that
for any $t,R$, $\lowerpotMdk(t,\R)$ increases with $k$.

To prove this claim we strengther the condition $\finalPotR \in \SP{2}$ used above to $\finalPotR \in \sp{4}$. In words, we assume that the function $\finalPotR(\R)$ is continuous and strictly positive and it's first four derivatives are continuous and strictly positive.

We use the sequence of discrete adversarial strategies
$\adversMdk, k=1,2,\ldots$ defined in
Section~\ref{sec:disc-game-strategies}.

\begin{theorem}\label{thm:smallerSteps}
  If $ \finalPotR \in \SP{4}$, and $\realT>0$  
  then for any $k>0$, any $t \in [0,\realT]$ and any $\R$
  $$\lowerpotMdkpar{k+1}(t,\R) >  \lowerpotMdkpar{k}(t,\R)$$
\end{theorem}

The proof of the theorem relies on a reduction to a simpler case:
dividing a single time step of duration $\tau$ into four time steps of duration  $\tau/4$

\begin{lemma} \label{lemma:half-step}
  If $ \finalPotTau \in \SP4$, and $\tau>0$ then for any $\R$
  $$\lowerpotMdkpar{1}(0,\R) >  \lowerpotMdkpar{0}(0,\R)$$
\end{lemma}
  
\proof The step size is $s_k=2^{-k}\sqrt{\tau}$, therefor
$s_0=\sqrt{\tau}, s_1=\frac{\sqrt{\tau}}{2}$. The time icrement is
$\Delta t_k = s_k^2$, therefor
$\Delta t_0=\tau, \Delta t_1=\frac{\tau}{4}$. In other words,
$k=0$ corresponds to a single step of size , while $k=1$ corresponds
to four steps of size $\frac{1}{2}$.

By definition $\finalPotTau(R)=\lowerpotMdkpar{0}(\tau,\R)=\lowerpotMdkpar{1}(\tau,\R)$

For $k=0$ we we get the recursion
\begin{equation}  \label{eqn:pot-recursion-0}
  \lowerpotMdkpar{0}(0, \R) =
  \frac{\lowerpotMdkpar{0}(\tau,\R-\sqrt{\tau})+
    \lowerpotMdkpar{0}(\tau,\R+\sqrt{\tau})}{2}
  =   \frac{\finalPotTau(\R-\sqrt{\tau})+
    \finalPotTau(\R+\sqrt{\tau})}{2}
\end{equation}

For $k=1$ we we have for $i=0,1,2,3$:
\begin{equation}   \label{eqn:lowerpotquarterstep}
   \lowerpotMdkpar{0}\paren{\frac{i}{4}\tau,\R}=
 \frac{\lowerpotMdkpar{0}\paren{\frac{i+1}{4}\tau,\R-\frac{1}{2}\sqrt{\tau}}+
   \lowerpotMdkpar{0}\paren{\frac{i+1}{4}\tau,\R+\frac{1}{2}\sqrt{\tau}}}{2}
 \end{equation}

\newcommand{\iter}[1]{\lowerpotMdkpar{1}\paren{\tau,\R {#1} \sqrt{\tau}}}
\newcommand{\iterzero}{\lowerpotMdkpar{1}\paren{\tau,\R}}

\newcommand{\fIter}[1]{\finalPotTau \paren{R {#1} \sqrt{\tau}}}
\newcommand{\fIterzero}{\finalPotTau \paren{R}}

\newcommand{\gIter}[1]{\finalPotTau \paren{R {#1} a}}
\newcommand{\gIterzero}{\finalPotTau \paren{R}}

Combining Equation~(\ref{eqn:lowerpotquarterstep}) for $k=0,1,2,3$ we get
\small
\begin{eqnarray} \label{eqn:pot-recursion-1}
  \lowerpotMdkpar{1}(0, \R)
  &=& \frac{1}{16}
      \left[\iter{-2}+4\iter{-} \right. \\
  &&+ \left. 6\iterzero +4\iter{+}+\iter{+2}\right] \nonumber\\
  &=& \frac{1}{16}
      \left[\fIter{-2}+4\fIter{-}+6\fIterzero +4\fIter{+}+\fIter{+2}\right] \nonumber
\end{eqnarray}
\normalsize the difference between
Equations~(\ref{eqn:pot-recursion-1}) and~(\ref{eqn:pot-recursion-0})
is \small
\begin{eqnarray} \label{eqn:pot-recursion-diff}
  \lefteqn{\lowerpotMdkpar{1}(0, \R) - \lowerpotMdkpar{0}(0, \R)}\\
&=&  \frac{1}{16}
\left[\fIter{-2}-4\fIter{-}+6\fIterzero -4\fIter{+}+\fIter{+2}\right] \nonumber
\end{eqnarray}
\normalsize Our goal is to show that the LHS of
Eqn.~\ref{eqn:pot-recursion-diff} is positive. This is equivalent to
proving positivity of
\begin{eqnarray}
g_a(\R) &=& \frac{2}{3a^2}\paren{\lowerpotMdkpar{1}(0, \R) -
  \lowerpotMdkpar{0}(0, \R)}\nonumber \\
& = &
\frac{1}{24 a^4}
      \left[\gIter{-2}-4\gIter{-}+6\gIterzero -4\gIter{+}+\gIter{+2}\right]
       \label{eqn:recursion-as-difference}
\end{eqnarray}
where $a=2 s_1=\sqrt{\tau}$

The function $g_a(\R)$ has a special form called ``divided
differences''. The proof of the following lemma uses this fact to show that 
Eqn~(\ref{eqn:recursion-as-difference}) is strictly postive.
\begin{lemma} \label{lemma:divdiff}
If $\finalPotTau \in \SP{4}$ and $\tau>0$, then $\forall \R, g_s(\R)>0$
\end{lemma}
The proof of Lemma~\ref{lemma:divdiff} is given in appendix~\ref{sec:divdiff}

\proof  of Theorem~\ref{thm:smallerSteps} \\
The proof is by double induction over $k$ and over $t_i$.
For $k=1,2,\ldots$ we consider the the loss step $s=2^{-k-1}\sqrt{\realT}$ and the time step 
$\Delta t = s^2=2^{-2k-2} \realT$. For each game iteration $i=0,\ldots,2^{2k}-1$
we fix the potential at time $t_1=(i+1) 2^{-2k}\realT$ and 
We consider the difference between the potential at $t_0=(i+1) 2^{-2k}\realT$

we take a finite backward induction over
$t=T-2^{-2k},T-2 \times 2^{-2k},T-3 \times 2^{-2k},\cdots,0$.
Our inductive claims are that $\pot_{k+1}(t,\R) > \pot_{k}(t,\R)$ and
$\pot_{k+1}(t,\R)$,$\pot_{k}(t,\R)$ are continuous, strongly convex and
have a strongly positive fourth derivative. That these claims carry over
from $t=T-i \times 2^{-2k}$ to  $t=T-(i+1) \times 2^{-2k}$ follows
directly from Lemma~\ref{lemma:n-strictly-convex}.

The theorem follows by forward induction on $k$.

\qed

Theorem~\ref{thm:seq-of-adv-strategies} characterizes the limit
\begin{equation} \label{eqn:k-limit}
  \lim_{k \to \infty} \upperpotMdk(t,\R_0) =
  \lim_{k \to \infty}\lowerpotMdk(t,\R_0)=
  {\cal N}(\R_0,\realT-t) \odot \finalPotR
\end{equation}

Theorem~\ref{thm:smallerSteps} states that increasing $k$ is always advantageous to the adversary.

Together these theorems show that the the min/max optimal potential function is $ {\cal N}(\R_0,\realT-t) \odot \finalPotR$.

\section{Brownian motion}
\label{sec:continuous}

There seems to be a paradox: the adversary prefers to set $s_i>0$ as
small as possible. On the other hand, there is no minimal strictly
positive number, so whatever the adversary chooses has to be
suboptimal. In other words, time is not continuous, it increases in
discrete steps. As that is the case, why is brownian motion still
the correct way to compute the potential?

One can use the following argument: the learner knows the range
$[-s_i,+s_i]$ for the next instantanous losses before it has to choose
the weights he places on the actions. On the other hand, it does not
know the range of the following losses, but he knows that the
adversary always prefers small ranges. The safe thing for the learner
to do is to assume that the following steps will be infinitesmaly
small, i.e. that the future losses form a brownian process

It is well known that the limit of random walks where $s \to 0$ and
$\deltat=s^2$ is the the Brownian or Wiener process
(see~\cite{kac1947random}).

An alternative characterization of Brownian Process is
$$ \P{}{X_{t+\deltat}=x_1 | X_t=x_0}=e^{-\frac{(x_1-x_0)^2}{2 \deltat}}$$

The backwards recursion that
defines the value function is the celebrated Backwards Kolmogorov
Equation with no drift and unit variance
\begin{equation} \label{eqn:Kolmogorov}
  \frac{\partial}{\partial t} \pot(t,\R)
  + \frac{1}{2} \frac{\partial^2}{\partial \R^2} \pot(t,\R)=0
\end{equation}
Given a final value function with a strictly positive fourth
derivative we can use Equation~(\ref{eqn:Kolmogorov}) to compute the
value function for all $0 \leq t \leq T$. We will do so in he next section.

\section{The continuous time game and bounds for easy
  sequences} \label{sec:easy}

In Section~\ref{sec:discrete} we have shown that the integer time game
has a natural extension to a setting where $\deltat_i = s_i^2$. We
also demonstrated sequences of adversarial strategies $S_1,S_2,\ldots$
such that $\sup_{k \to \infty} {\lowerpot}_k(0,\R) = $

We characterized the optimal adversarial strategy for the discrete
time game (Section~\ref{sec:discrete-Time-Game}), which corresponds
to the adversary choosing the loss to be $s_i$ or $-s_i$ with equal
probabilities. A natural question at this point is to characterize the
regret when the adversary is not optimal, or the sequences are ``easy''.

To see that such an improvement is possible, consider the following
{\em constant} adversary. This adversary associates the same loss to
all actions on iteration $i$, formally, $\adversM(i,\R) = l$. In this
case the average loss is also equal to $l$, $\ell(i)=l$ which means
that all of the instantaneous regrets are $r=l-\ell(t_i) = 0$, which,
in turn, implies that $\state(i) = \state(i+1)$. As the state did not
change, it makes sense to set $t_{i+1}=t_i$, rather than
$t_{i+1}=t_i+s_i^2$.

We observe two extremes for the adversarial behavior. The constant
adversary described above for which $t_{i+1} = t_i$, and the random walk adversary described
earlier, in which each action is split into two, one half with loss
$-s_i$ and the other with loss $+s_i$. In which case $t_{i+1} =
t_i+s_i^2$ which is the maximal increase in $t$ that the adversary can
guarantee. The analysis below shows that these are two extremes on a
spectrum and that intermediate cases can be characterized using a
variance-like quantity.

We define a variant of the discrete time game
(\ref{sec:discrete-Time-Game}) For concreteness we include the
learner's strategy, which is the limit of the strategy in the discrete
game when $s_i \to 0$.

\begin{figure}[ht!]
\framebox{
\begin{minipage}[t]{6.4in}
Set $t_1=0$ \\
Fix maximal step $0<s<1$ \\
On iteration $i=1,2,\ldots$

\begin{enumerate}
\item  If $t_i=T$ the game terminates.
\item Given $t_{i}$, the learner chooses a distribution
  $\learnerM(i)$ over $\reals$:
  \begin{equation} \label{eqn:learner-strat-cc}
  \learnerM^{cc}(t,\R) =  \frac{1}{Z^{cc}}
  \left. \frac{\partial}{\partial r} \right|_{r=\R} \pot(t,r)
  \mbox{ where } Z^{cc} = \E{\R \sim \state(t_i)}{\left. \frac{\partial}{\partial r} \right|_{r=\R} \pot(t,r)}
\end{equation}

\item The adversary chooses a {\em step size} $0<s_i\leq s$ and a mapping from $\reals$ to distributions
  over $[-s_i,+s_i]$: $\adversM(t): \reals \to \Delta^{[-s_i,+s_i]}$
\item The aggregate loss is calculated:
  \begin{equation} 
    \ell(t_i)=\E{\R \sim \state(t_i)}{\learnerM^{cc}(t_i,\R)
      \Bias(t_i,\R)},\;\mbox{ where } \Bias(t_i,\R) \doteq \E{y \sim \adversM(t_i,\R)}{y}
  \end{equation}
  the aggregate loss is restricted to $|\ell(t_i)| \leq c s_i^2$.
\item  Increment $t_{i+1} = t_{i} + \deltat_i$ where
\begin{equation} \label{eqn:deltat}
  \deltat_i=
  \E{\R \sim \state(t_i)}{H(t_i,\R) \;\; \E{y \sim \adversM(t_i,\R)}{(y-\ell(t_i))^2}}
\end{equation}
Where
\begin{equation}
 H(t_i,R)=\frac{1}{Z^H} \left. \frac{\partial^2}{\partial r^2} \right|_{r=\R} \pot(t_i,r)
  \mbox{ and } Z^H = \E{\R \sim \state(t_i)}{\left. \frac{\partial^2}{\partial r^2} \right|_{r=\R} \pot(t_i,r)}
\end{equation}

\item The state is updated.
  $$\state(t_{i+1}) = \E{\R \sim \state(t_{i})}{\adversM(t_i)(\R)\oplus (\R-\ell(t_i))}
  $$
\end{enumerate}
\end{minipage}}
\caption{The continuous time game and learner strategy\label{sec:contin-Time-Game}}
\end{figure}

Our characterization applies to the limit where the $s_i$ are small. Formally, we define
\begin{definition}
We say that an instance of the discrete time game is
$(n,s,\tau)$-bounded if it consists of $n$ iterations and $\forall\;\; 0<i\leq n,\;\; s_i < s$ and $\sum_{j=1}^n s_j^2=\tau$
\end{definition}

Note that $\tau>t_n$ and that $\tau$ depends only on the ranges $s_i$
while $t_n$ depends on the variance. $t_n = T$ 
is the dominant term in the regret bound, while $\tau$ controls the
error term.

\newpage

\begin{theorem} \label{thm:variancebound} Let $\pot \in \SP{\infty}$
  be a potential function that satisfies the Kolmogorov backward
  equation~(\ref{eqn:Kolmogorov}).
  Fix the total time $\tau$ and let $G_n$ be an $(n,
  \sqrt{\frac{\tau}{n}},\tau)$-bounded game. Let $n \to \infty$.

Then 
$$\score(\state(\tau)) \leq \score(\state(0))+O\paren{\frac{1}{\sqrt{n}}}$$
\end{theorem}

The proof is given in appendix~\ref{appendix:ProofOfVarianceBound}

If we define

\begin{equation} \label{eqn:Vn}
  V_n = t_n = \sum_{i=1}^n \deltat_i= 
  \sum_{i=1}^n \E{\R \sim \state(t_i)}{\E{y \sim \adversM(t_i,\R)}{H(t_i,\R) ((y-\ell(t_i))^2)}}
\end{equation}

We can use $V_n$ instead of $T$ giving us a variance based bound.


\section{Anytime potential functions}

The results up to this point hold for any potential function in
$\SP{4}$. Given a final potential function $\finalPotR \in \SP{4}$ we
can compute the potential for any $0 \leq t \leq \realT$ and any $R$ using the equation 
\begin{equation} \label{eqn:convol-with-normal}
\pot(t,\R)={\cal N}(\R_0,\realT-t) \odot \finalPotR
\end{equation}
By using uses the $\R$-derivative of this potential function to define
the weights the learner guarantees that the final average score is at
most $\pot(0,0)$.

A major limition of this result is that the horizon $\realT$ is set in
advance.  It is desirable that the potential is defined without
knowledge of the horizon.  In what follows we show that Hedge and
NormalHedge can both be used in such ``anytime'' algorithms.

Our solution is based on the observation that a potential function satisfies Eqn~(\ref{eqn:convol-with-normal}) if and only if it satisfies 
the Kolmogorov backwards PDE~(\ref{eqn:Kolmogorov}):
\begin{equation} 
  \frac{\partial}{\partial t} \pot(t,\R)
  + \frac{1}{2} \frac{\partial^2}{\partial r^2} \pot(t,\R)=0
\end{equation}
The potential $\finalPotR \in \SP{4}$ defines a boundary condition of the PDE.

We derive our anytime algorithm by finding solutions to the Kolmogorov
PDE that are not restricted in time, and that have a fixed parametric form.
In other words, the evolution of the potential with time is defined by changing the parameter values, without changing the form.

We consider two solutions to the PDE, the exponential potential and
the NormalHedge potential. We give the form of the potential function
that satisfies Kolmogorov Equation~\ref{eqn:Kolmogorov}, and derive
the regret bound corresponding to it.

{\bf The exponential potential function} which corresponds to exponential
  weights algorithm corresponds to the following equation
\[
    \pot_{\mbox{\tiny exp}}(\R,t) = e^{\sqrt{2} \eta \R - \eta^2 t}
  \]
  Where $\eta>0$ is the learning rate parameter.
  
Given $\epsilon$ we choose $\eta = \sqrt{\frac{\ln (1/\epsilon)}{t}}$
we get the regret bound that holds for any $t>0$
  \begin{equation}
    \R_\epsilon \leq \sqrt{2 t \ln \frac{1}{\epsilon}}
  \end{equation}
Note that the algorithm depends on the choice of $\epsilon$, in other
words, the bound does {\em not} hold for all values of $\epsilon$ at
the same time.

{\bf The NormalHedge value} is
\begin{equation} \label{eqn:NormalHedge}
  \pot_{\mbox{\tiny NH}}(\R,t) = \begin{cases}
    \frac{1}{\sqrt{t+1}}\exp\left(\frac{\R^2}{2(t+1)}\right)
    & \mbox{if } \R \geq 0  \\
  \frac{1}{\sqrt{t+1}} & \mbox{if } \R <0
  \end{cases}
\end{equation}
The function $\pot_{\mbox{NH}}(\R,t)$,
restricted to $\R\geq 0$ is in $\SP{4}$ and is a constant for $\R \leq 0$.

The regret bound we get is:
\begin{equation}
  \R_\epsilon \leq \sqrt{(t+1) \left(2 \ln \frac{1}{2\epsilon}+
      \ln (t+1)\right)}
\end{equation}
This bound is slightly larger than the bound for exponential weights,
however, the NormalHedge bound holds simultaneously for all
$\epsilon>0$ and the algorithm requires no tuning.

\bibliographystyle{plain}
\bibliography{ref.bib,bib.bib}

\appendix
\section{Proof of Theorem~\ref{thm:simulBoundAveragePot} \label{proof:simulBoundAveragePot}}
\proof
  \begin{itemize}
  \item
  {\bf $\state$ satisfies a simultaneous bound for $B$ if it satisfies an
  average potential bound for $\pot = B^{-1}$}\\
Assume by contradiction that $\state$ does not satisfy the simultaneous bound. In
other words there exists $a \in \reals$ such that
$\P{\R \sim \state}{\R > a} > B(a)$. From Markov inequality and the fact
that $\phi$ is non decreasing we get
\[
  \E{\R \sim \state}{\pot(\R)} \geq \phi(a) \P{\R \sim \state}{\R > a} >
  \phi(a) B(a) = \frac{B(a)}{B(a)}=1
\]
but $ \E{\R \sim \state}{\pot(\R)} >1$ contradicts the average potential
assumption for the potential $\phi(\R) = B(\R)^{-1}$
\item
{\bf $\state$ satisfies an
  average potential bound for $\pot = B^{-1}$ if it satisfies a simultaneous bound for $B$}\\
As $\phi$ is a non-decreasing function, and assuming $\R,\R'$ are drawn
independently at random according to $\state$:
\begin{eqnarray}
  \E{\R \sim \state}{\pot(\R)} & = & \E{\R \sim \state}{\pot(\R)
                                  \P{\R' \sim \state}{\phi(R') \geq \phi(\R)}} \\
                            & \leq & \E{\R \sim \state}{\pot(\R)
                                     \P{\R' \sim \state}{R' \geq \R}} \\
                            & < & \E{\R \sim \state}{\pot(\R) B(\R)} \\
                            & = & \E{\R \sim \state}{\frac{B(\R)}{B(\R)}}
                                  = \E{\R \sim \state}{1} = 1
\end{eqnarray}
\end{itemize}
\qed

\section{Divided differences of a function} \label{sec:divdiff}

The function $g_s(\R)$ has a special form called ``divided difference''
that has been extensively studied ~\cite{popoviciu1965certaines,butt2016generalization, de2005divided}.
and is closely related to to derivatives of different orders. Using
this connection and the fact that $\pot(\cdot,\R) \in \SP{4}$ we prove
the following lemma:

We conclude that if $\pot(t',\R)$ has a strictly positive fourth
derivative then $\pot_{k+1}(t,\R) > \pot_{k}(t,\R)$ for all $\R$, proving
the first part of the lemma.

The second part of the lemma follows from the fact that
both $\pot_{k+1}(t,\R)$ and $\pot_{k}(t,\R)$ are convex combinations of
$\pot(t,\R)$ and therefor retain their continuity and convexity properties.

A function $\finalPot{}$ that satisfies
inequality~\ref{eqn:4thOrderConvex} is said to be {\em 4'th order convex}
(see details in in~\cite{butt2016generalization}).

Following\cite{butt2016generalization} we give a brief review of
divided differences and of $n$-convexity.

Let $f:[a,b] \to \reals$ be a function from the segment $[a,b]$ to the
reals.

\begin{definition}[$n$'th order divided difference of a function]
  The $n$'th order divided different of a function $f:[a,b] \to
  \reals$ at mutually distinct and ordered points $a \leq x_0 < x_1
  < \cdots < x_n \leq b$
  defined recursively by
  \[ [x_i; f] = f(x_i), \; i \in 0,\ldots n,\]
  \[ [x_0,\ldots,x_n;f] =
    \frac{[x_1,\ldots,x_n;f]-[x_0,\ldots,x_{n-1};f]}{x_n-x_0} \]
\end{definition}

\begin{definition}[$n$-convexity]
 A function $f:[a,b] \to \reals$ is said to be $n$-convex  $n \geq 0$
 if and only if for all choices of $n+1$ distinct points: $a \leq x_0 < x_1
  < \cdots < x_n \leq b$, $[x_0,\ldots,x_n;f]\geq 0$ holds.
\end{definition}
$n$-convexity is has a close connection to the sign of $f^{(n)}$ - the $n$'th
derivative of $f$, this connection was proved in 1965 by
popoviciu~\cite{popoviciu1965certaines}.
\begin{theorem} \label{thm:popo}
If $f^{(n)}$ exists then f is $n$-convex if and only if $f^{(n)}\geq 0$.
\end{theorem}

The next lemma states that the function $g(\R)>0$ as defined in
Equation~(\ref{eqn:divdiff}).

\proof {\bf of Lemma~(\ref{lemma:divdiff})

Fix $t$ and define $f(x) = \pot(t,x)$.
Let $(x_0,x_1,x_2,x_3,x_4)=(\R-2 s,\R-s,\R,\R+s,\R+2s)$

Using this notation we can rewrite $g(\R)$ in the form
\begin{equation}
  h(x_0,x_1,x_2,x_3,x_4) =  \frac{1}{24s^4} \paren{f(x_4)- 4f(x_3)+ 6f(x_2)-
    4f(x_1)+ f(x_0)}
\end{equation}
Is the 4-th order divided difference of $\pot(t,\cdot)$

\begin{enumerate}
\item
$$[x_i;f] = f(x_i)$$
\item
  $$[x_i,x_{i+1};f]=\frac{f(x_{i+1})-f(x_i)}{s}$$
\item
  $$[x_i,x_{i+1},x_{i+2};f] =
  \frac{\frac{f(x_{i+2})-f(x_{i+1})}{s}-\frac{f(x_{i+1})-f(x_i)}{s}}{2s}
  =\frac{f(x_{i+2})-2f(x_{i+1})+f(x_i)}{2 s^2}
  $$
\item
  \begin{eqnarray*}
    [x_i,x_{i+1},x_{i+2},x_{i+3};f]& = &
    \frac{\frac{f(x_{i+3})-2f(x_{i+2})+f(x_{i+1})}{2
    s^2}-\frac{f(x_{i+2})-2f(x_{i+1})+f(x_i)}{2
    s^2}}{3s}\\
    &=& \frac{f(x_{i+3}) -3f(x_{i+2})+3f(x_{i+1})-f(x_i)   }{6 s^3}
  \end{eqnarray*}
\item
  \begin{eqnarray*}
    [x_i,x_{i+1},x_{i+2},x_{i+3},x_{i+4};f]& = &
    \frac{\frac{f(x_{i+4}) -3f(x_{i+3})+3f(x_{i+2})-f(x_{i+1})   }{6 s^3}
    - \frac{f(x_{i+3}) -3f(x_{i+2})+3f(x_{i+1})-f(x_i)   }{6 s^3}}
    {4s}\\
    &=& \frac{f(x_{i+4})-4f(x_{i+3})+6f(x_{i+2})-4f(x_{i+1})+f(x_i)}{24s^4}
  \end{eqnarray*}
\end{enumerate}

\qed
\section{Proof of Theorem~\ref{thm:variancebound}}
\label{appendix:ProofOfVarianceBound}
We start with two technical lemmas
\begin{lemma} \label{lemma:infiniteexpectations}
Let $f(x) \in \SP{2}$, i.e. $f(x), f'(x),f''(x) >0$ for all $x \in
\reals$, let $h(x)$ be a uniformly bounded function: $\forall x,\;\; |h(x)|<1$.
Let $\state$ be a distribution over $\reals$.
If $\E{x \sim \state}{f(x)}$ is well-defined (and finite) , then 
$\E{x \sim \state}{h(x) f'(x)}$ is well defined (and finite) as well.
\end{lemma}
\proof
Assume by contradiction that $\E{x \sim \state}{h(x) f'(x)}$ is
undefined. Define $h^+(x) = \max(0,h(x))$.
As $f'(x)>0$, this implies that either $\E{x \sim \state}{h^+(x)
  f'(x)}=\infty$ or $\E{x \sim \state}{(-h)^+(x) f'(x)}=\infty$ (or both). 

Assume wlog that $\E{x \sim \state}{h^+(x) f'(x)}=\infty$. As
$f'(x)>0$ and $0 \leq h^+(x) \leq 1$ we get that $\E{x \sim
  \state}{f'(x)}=\infty$.
As $f(x+1) \geq f'(x)$ we get that $\E{x \sim
  \state}{f(x)}=\infty$ which is a contradiction.
\qed

\newcommand{\Dx}{\Delta x}
\newcommand{\Dy}{\Delta y}
\begin{lemma} \label{lemma:Taylor2D}
Let $f(x,y)$ be a differentiable function with continuous derivatives
up to degree three. Then
\begin{eqnarray}
  &&f(x_0+\Dx,y_0+\Dy) = f(x_0,y_0)
  + \atI{\frac{\partial}{\partial x}} \Dx 
  + \atI{\frac{\partial}{\partial y}} \Dy \\
  &+&\frac{1}{2} \atI{\frac{\partial^2}{\partial x^2}} \Dx^2
      +\atI{\frac{\partial^2}{\partial x\partial y}} \Dx\Dy
      +\frac{1}{2} \atI{\frac{\partial^2}{\partial y^2}} \Dy^2\\
  &+&\frac{1}{6} \atII{\frac{\partial^3}{\partial x^3}} \Dx^3
      +\frac{1}{2} \atII{\frac{\partial^3}{\partial x^2 \partial y}} \Dx^2\Dy\\
  &&+ \frac{1}{2} \atII{\frac{\partial^3}{\partial x \partial y^2}} \Dx\Dy^2
    + \frac{1}{6} \atII{\frac{\partial^3}{\partial y^3}} \Dy^3
\end{eqnarray}
for some $0\leq t \leq 1$.
\end{lemma}
\proof {\em of Lemma~\ref{lemma:Taylor2D}} 
Let $F:[0,1] \to \reals$ be defined as  $F(t)=f(x(t),y(t))$ where
$x(t) = x_0+t\Dx$ and $y(t)=y_0+t\Dy$. Then $F(0)=f(x_0,y_0)$ and
$F(1)=f(x_0+\Dx,y_0+\Dy)$. It is easy to verify that
$$ \frac{d}{dt}F(t)
=\frac{\partial}{\partial x} f(x(t),y(t))\Dx
+ \frac{\partial}{\partial y} f(x(t),y(t))\Dy
$$
and that in general:
\begin{equation} \label{eqn:d.dn.F}
\frac{d^n}{d t^n} F(t) = \sum_{m=1}^n {n \choose m}
\frac{\partial^n}{\partial x^m \partial y^{n-m}} f(x_0+t \Dx,y_0+t\Dy)
\Dx^m \Dy^{n-m}
\end{equation}
As $f$ has partial derivatives up to degree 3, so does $F$. Using the
Taylor expansion of $F$ and the intermediate point theorem we get that
\begin{equation} \label{eqn:Taylor.F}
  f(x_0+\Dx,y_0+\Dy) = F(1) = F(0)+\frac{d}{dt}F(0)
  +\frac{1}{2}\frac{d^2}{dt^2}F(0)
  +\frac{1}{6}\frac{d^3}{dt^3}F(t')
\end{equation}
Where $0 \leq t' \leq 1$. Using Eq~(\ref{eqn:d.dn.F}) to expand each
term in Eq.~(\ref{eqn:Taylor.F}) completes the proof.
\qed

\proof {\em of Theorem~\ref{thm:variancebound}}\\
We prove the claim by an upper bound on the increase of potential that holds for any iteration $1 \leq i \leq n$:
\begin{equation} \label{proof:onestep}
\score(\state(t_{i+1})) \leq \score(\state(t_i)) + a s_i^3 \mbox{ for some constant } a>0
\end{equation}
Summing inequality~(\ref{proof:onestep}) over all iterations we get that 
\begin{equation} \label{proof:allsteps}
\score(\state(T)) \leq \score(\state(0)) + c \sum_{i=1}^n s_i^3 \leq 
\score(\state(0)) + a s \sum_{i=1}^n s_i^2 = 
\score(\state(0)) + a s T
\end{equation}
From which the statement of the theorem follows.

We now prove inequality~(\ref{proof:onestep}). 
We use the notation $r=y -\ell(i)$ to denote the instantaneous regret at iteration $i$.

Applying Lemma~\ref{lemma:Taylor2D} to
$\pot(t_{i+1},\R_{i+1})=\pot(t_i+\deltat_i,\R_i+r_i)$  we get
\begin{eqnarray} 
    \pot(t_i+\deltat_i,\R_i+r_i) & =&  
    \pot(t_i,\R_i)\\
    &+&\at{\frac{\partial}{\partial \rho}} r_i \\
    &+&\at{\frac{\partial}{\partial \tau}}  \deltat_i \\
    &+& \frac{1}{2} \at{\frac{\partial^2}{\partial \rho^2}} r_i^2 \\
    &+& \at{\frac{\partial^2}{\partial r \partial \tau}} r_i \deltat_i \label{term:Taylor_rdt}\\
    &+& \frac{1}{2} \at{\frac{\partial^2}{\partial \tau^2}} \deltat_i^2 \label{term:Taylor_dtsquare}\\
    &+& \frac{1}{6} \att{\frac{\partial^3}{\partial \rho^3}} r_i^3 \label{term:Taylor_r3}\\
    &+& \frac{1}{2} \att{\frac{\partial^3}{\partial \rho^2 \partial \tau}} r_i^2\deltat_i \label{term:Taylor_r2t}\\
    &+& \frac{1}{2} \att{\frac{\partial^3}{\partial \rho \partial \tau^2}} r_i\deltat_i^2 \label{term:Taylor_rt2}\\
    &+& \frac{1}{6} \att{\frac{\partial^3}{\partial \tau^3}} \deltat_i^3 \label{term:Taylor_t3}
\end{eqnarray}
for some $0 \leq g \leq 1$.

By assumption $\pot$ satisfies the Kolmogorov backward equation:
\begin{equation*} 
  \frac{\partial}{\partial \tau} \pot(\tau,\rho)
  = -\frac{1}{2} \frac{\partial^2}{\partial r^2} \pot(\tau,\rho)
\end{equation*}
Combining this equation with the exchangeability of the order of
partial derivative (Clairiaut's Theorem) we can substitute all
partial derivatives with respect to $\tau$ with partial derivatives
with respect to $\rho$ using the following equation.
\[
  \frac{\partial^{n+m}}{\partial \rho^n \partial \tau^m} \pot(\tau,\rho)=
  (-1)^m \frac{\partial^{n+2m}}{\partial \rho^{n+2m}} \pot(\tau,\rho)
\]
Which yields
\begin{eqnarray}
      \pot(t_i+\deltat_i,\R_i+r_i) & =&  
    \pot(t_i,\R_i)\\
    &+& \at{\frac{\partial}{\partial \rho}} r_i \label{term:coll1}\\
    &+& \at{\frac{\partial^2}{\partial \rho^2}} \paren{\frac{r_i^2}{2}-\deltat_i} \label{term:coll2}\\
    &-& \at{\frac{\partial^3}{\partial \rho^3}} r_i \deltat_i \label{term:coll3}\\
    &+& \frac{1}{2} \at{\frac{\partial^4}{\partial \rho^4}} \deltat_i^2 \label{term:coll4}\\
    &+& \frac{1}{6} \att{\frac{\partial^3}{\partial \rho^3}} r_i^3 \label{term:coll5}\\
    &-& \frac{1}{2} \att{\frac{\partial^4}{\partial \rho^4}} r_i^2\deltat_i \label{term:coll6}\\
    &+& \frac{1}{2} \att{\frac{\partial^5}{\partial \rho^5}} r_i\deltat_i^2 \label{term:coll7}\\
    &-& \frac{1}{6} \att{\frac{\partial^6}{\partial \rho^6}} \deltat_i^3 \label{term:coll8}
\end{eqnarray}

  From the assumption that the game is $(n,s,T)$-bounded we get that 
  \begin{enumerate}
  \item $|r_i| \leq s_i +c s_i^2 \leq 2 s_i$
  \item $\deltat_i \leq s_i^2 \leq s^2$
  \end{enumerate}

  given these inequalities we can rewrite the second factor in each
  term as follows, where $|h_i(\cdot)|\leq 1$
  \begin{itemize}
  \item {\bf For~(\ref{term:coll1}):}
    $r_i=2s_i\frac{r_i}{2s_i}=2s_ih_1(r_i)$.
  \item {\bf For~(\ref{term:coll2}):}
    $r_i^2 - \frac{1}{2}\deltat_i = 4s_i^2\frac{r_i^2 -
      \frac{1}{2}\deltat_i}{4s_i^2} = 4s_i^2 h_2(r_i,\deltat_i)$
  \item {\bf For~(\ref{term:coll3}):} $r_i \deltat_i = 2s_i^3
    \frac{r_i \deltat_i}{2s_i^3} = 2s_i^3 h_3(r_i,\deltat_i)$
  \item {\bf For~(\ref{term:coll4}):} $\deltat_i^2 =
    s_i^4\frac{\deltat_i^2}{s_i^4} = s_i^3 h_4(\deltat_i)$
  \item {\bf For~(\ref{term:coll5}):} $r_i^3 = 8s_i^3
    \frac{r_i^3}{8s_i^3} = 8s_i^3 h_5(r_i,\deltat_i)$
  \item {\bf For~(\ref{term:coll6}):} $r_i^2 \deltat_i = 4s_i^4
    \frac{r_i^2 \deltat_i}{4s_i^4} = 4s_i^3 h_6(r_i,\deltat_i)$
  \item {\bf For~(\ref{term:coll7}):} $r_i \deltat_i^2 = 2s_i^5
    \frac{r_i \deltat_i^2}{2s_i^5}$
  \item {\bf For~(\ref{term:coll8}):} $\deltat_i^3 = s_i^6 \frac{\deltat_i^3}{s_i^6}$
\end{itemize}
  We therefor get the simplified equation
  
  \begin{eqnarray*} 
     \pot(t_i+\deltat_i,\R_i+r_i) & =&  \pot(t,\R)+\at{\frac{\partial}{\partial r}} r
    + \at{\frac{\partial}{\partial t}} \deltat \\
                                  &+& 
                                      \frac{1}{2}  \at{\frac{\partial^2}{\partial r^2}} r^2\\
                                  &+& \at{\frac{\partial^2}{\partial r \partial t}} r_i \deltat_i \label{term:Taylor_collected_rdt}\\
                                  &+& \frac{1}{6} \at{\frac{\partial^3}{\partial r^3}} r_i^3 \label{term:Taylor_collected_r3}
                                      + O(s^4)
\end{eqnarray*}

and therefor
  \begin{eqnarray} 
     \pot(t_i+\deltat_i,\R+r) &=& \pot(t_i,\R) +
                                  \at{\frac{\partial}{\partial r}} r
                                  \nonumber \\
    &+& \at{\frac{\partial^2}{\partial r^2}} (r^2 - \deltat_i) +
        O(s^3) \label{eqn:Taylor}
\end{eqnarray}

Our next step is to consider the expected value of~(\ref{eqn:Taylor}) wrt $\R \sim \state(t_i)$,
$y \sim \adversM(t_i,\R)$ for an arbitrary adversarial strategy
$\adversM$.

We will show that the expected potential does not increase:
\begin{equation} \label{eqn:deltatislargeenough}
     \E{\R \sim \state(t_i)}{ \E{y \sim \adversM(t_i,\R)}{\pot(t_i+\deltat_i,\R+y-\ell(t_i))}} \leq \E{\R \sim \state(t_i)}{\pot(t_i,\R)}
\end{equation}

Plugging Eq~(\ref{eqn:Taylor}) into the LHS of
Eq~(\ref{eqn:deltatislargeenough}) we get
\begin{eqnarray}
  \lefteqn{\E{\R \sim \state(t_i)}{\E{y \sim \adversM(t_i,\R)}{\pot(t_i+\deltat_i,\R+y-\ell(t_i))}}} \\
  &=& \E{\R \sim \state(t_i)}{\pot(t_i,\R)} \label{eqn:contin0}\\
  &+& \E{\R \sim \state(t_i)}{\E{y \sim \adversM(t_i,\R)}{\at{\frac{\partial}{\partial r}} (y-\ell(t_i))}} \label{eqn:contin1}\\
  &+& \E{\R \sim \state(t_i)}{\E{y \sim
      \adversM(t_i,\R)}{\at{\frac{\partial^2}{\partial r^2}}
      ((y-\ell(t_i))^2 - \deltat_i)}}
  \label{eqn:contin2}\\
  &+& O(s^3) \label{eqn:contin3}
\end{eqnarray}
Some care is needed here. we need to show that the expected value
are all finite. We assume that the expected potential
(Eq~(\ref{eqn:contin0}) is finite. Using
Lemma~\ref{lemma:infiniteexpectations} this implies that the expected
value of higher derivatives of $\frac{\partial}{\partial \R} \pot(\R)$
are also finite.\footnote{I need to clean this up and find an argument
  that the expected value for mixed derivatives is also finite.}

To prove inequality~(\ref{proof:onestep}), we need to show that the
terms~\ref{eqn:contin1} and \ref{eqn:contin2} are smaller or equal to
zero.
~\\~\\~\\
{\bf Term~(\ref{eqn:contin1}) is equal to zero:}\\
As $\ell(t_i)$ is a constant
relative to $\R$ and $y$, and $\at{\frac{\partial}{\partial r}}$ is a
constant with respect to $y$ we can rewrite~(\ref{eqn:contin1}) as
\begin{equation} \label{eqnterm1.1}
  \E{\R \sim \state(t_i)}{\at{\frac{\partial}{\partial r}}
    \E{y \sim \adversM(t_i,\R)}{y} }
- \ell(t_i) \E{\R \sim \state(t_i)}{\at{\frac{\partial}{\partial r}}}
\end{equation}

Combining the definitions of $\ell(t)$~(\ref{eqn:ell-discrete}) and~
and the learner's strategy
$\learnerM^{cc}$~(\ref{eqn:learner-strat-cc}) we get that
\begin{eqnarray}
\ell(t_i) &=& \E{\R \sim \state{t_i}}{\frac{1}{Z}
              \at{\frac{\partial}{\partial r}}
              \E{y \sim \adversM(i,\R)}{y}} \mbox{ where }
              Z=\E{\R \sim \state{t_i}}{\frac{1}{Z}
              \at{\frac{\partial}{\partial r}}}
              \label{eqnterm1.2}
\end{eqnarray}

Plugging~(\ref{eqnterm1.2}) into (\ref{eqnterm1.1}) and recalling the
requirement that $\ell(t_i)<\infty$ we find that
term~(\ref{eqn:contin1}) is equal to zero.

~\\~\\~\\
{\bf Term~(\ref{eqn:contin2}) is equal to zero:}\\
As $\deltat_i$ is a constant relative to $y$, we can take it
outside the expectation and plug in the definition of $\deltat_i$ (\ref{eqn:deltat})
\begin{equation} \label{eqn:term2.1}
  \E{\R \sim \state(t_i)}{\E{y \sim
      \adversM(t_i,\R)}{\adversM(t_i,\R)}{\at{\frac{\partial^2}{\partial r^2}}
      (y-\ell(t_i))^2} - \deltat_i}=
  \deltat_i - \deltat_i =0
\end{equation}
Where $G(t_i,\R)$ is defined in Equation~(\ref{eqn:SecondOrderDerivative})
We find that (\ref{eqn:contin2}) is zero.

Finally (\ref{eqn:contin3}) is negligible relative to the other terms
as $s \to 0$.
\qed

\end{document}